\pdfoutput=1
\documentclass{article}

\PassOptionsToPackage{numbers, compress}{natbib}

\usepackage[preprint]{neurips_2022}

\usepackage[utf8]{inputenc} 
\usepackage[T1]{fontenc}    
\usepackage{hyperref}       
\usepackage{url}            
\usepackage{booktabs}       
\usepackage{amsfonts}       
\usepackage{nicefrac}       
\usepackage{microtype}      
\usepackage{xcolor}         

\usepackage{amsmath}
\usepackage{amsthm}
\usepackage{amsfonts}
\usepackage{amssymb}
\usepackage{bm}

\usepackage[pdftex]{graphicx}
\graphicspath{ {./figures/} }
\usepackage{comment}
\usepackage{csquotes}
\theoremstyle{definition}
\newtheorem{definition}{Definition}[section]

\usepackage{setspace}
\usepackage{enumitem}

\DeclareMathOperator*{\argmin}{arg\,min}
\usepackage{algorithm, algorithmic}
\usepackage{hyperref}

\title{Defense Against Gradient Leakage Attacks \\via Learning to Obscure Data}

\author{%
    Yuxuan Wan \\
    Michigan State University \\
    \texttt{wanyuxua@msu.edu} \\
    \And
    Han Xu \\
    Michigan State University \\
    \texttt{xuhan1@msu.edu} \\
    \And
    Xiaorui Liu \\
    Michigan State University \\
    \texttt{xiaorui@msu.edu} \\
    \And
    Jie Ren \\
    Michigan State University \\
    \texttt{renjie3@msu.edu} \\
    \And
    Wenqi Fan\\
    The Hong Kong Polytechnic University\\
    \texttt{wenqifan03@gmail.com}\\
    \And
    Jiliang Tang\\
    Michigan State University\\
    \texttt{tangjili@msu.edu}
}

\begin{document}

\maketitle

\begin{abstract}
  Federated learning is considered as an effective privacy-preserving learning mechanism that separates the client's data and model training process. However, federated learning is still under the risk of privacy leakage because of the existence of attackers who deliberately conduct gradient leakage attacks to reconstruct the client data. Recently, popular strategies such as gradient perturbation and input encryption have been proposed to defend against gradient leakage attacks. Nevertheless, these defenses can either greatly sacrifice the model performance, or be evaded by more advanced attacks. In this paper, we propose a new defense method to protect the privacy of clients' data by learning to obscure data. Our defense method can generate synthetic samples that are totally distinct from the original samples, but they can also maximally preserve their predictive features and guarantee the model performance. Furthermore, our defense strategy makes the gradient leakage attack and its variants extremely difficult to reconstruct the client data. Through extensive experiments, we show that our proposed defense method obtains better privacy protection while preserving high accuracy compared with state-of-the-art methods.
\end{abstract}

\section{Introduction}

Machine learning models are playing an increasingly important role in applications which highly interact with human society. Meanwhile, there is also a strong demand to protect private and sensitive information in data during the model training process. For example, Convolutional Neural Networks~\cite{kalchbrenner2014convolutional, he2016deep} are widely adopted to diagnose diseases based on patients' medical images~\cite{castiglioni2021ai, shen2017deep}, and language models~\cite{lavrenko2017relevance, zhang2018deep} are widely used to help users to compose text messages and correct their grammar errors~\cite{sneiders2010automated,wang2020comprehensive}. Thus, protecting the privacy of clients’ data while training machine learning models with high performance becomes a critical mission to build trustworthy AI.

Recently, federated learning~\cite{shokri2015privacy,mcmahan2017communication, bonawitz2019towards, kairouz2019advances} has been considered as an efficient and effective privacy-preserving machine learning framework, which has greatly relieved people's concern about privacy leakage. In federated learning, the clients can keep their private data in their local devices without sharing their data with other parties. They only need to train the model in their local devices and share the model update or gradient with a central server. The server aggregates these updates to construct a global model and sends the new model parameters to the selected clients. In the process of federated learning, the data of each client is unseen to others, so private information is believed to be protected.

However, recent studies find that federated learning is still under the huge threat of privacy leakage. For instance, the works~\cite{zhu2020deep, geiping2020inverting, yin2021see} propose an attack method, namely Gradient Leakage Attack, to reconstruct the original private data sample, i.e., a client's image, by exploiting the shared gradient information or model updates in federated learning. To better protect privacy, recently multiple countermeasures have been proposed to defend against gradient leakage attacks. There are mainly two types of defense strategies. 
The first type is gradient perturbation~\cite{zhu2020deep, sun2021soteria}, which directly prunes the shared gradients in federated learning to defeat gradient leakage attacks. However, recent works~\cite{huang2021evaluating} find that it is usually required to prune too much gradient information to fully defeat gradient leakage attacks, which will greatly hurt the model accuracy. The second type is input data encryption~\cite{zhang2017mixup, huang2020instahide}, which encrypts the data and hides private information in client data. However, current state-of-the-art encryption methods~\cite{huang2020instahide} can also be evaded by adaptive attack methods ~\cite{carlini2020attack}. Therefore, an effective defense method which can reliably protect the privacy of clients while preserving model accuracy is still highly demanded.

In this work, we propose a novel input data encryption method, which aims to encrypt clients' data via eliminating the private features using an optimization approach. Specifically, our method is motivated by~\cite{ilyas2019adversarial, tsipras2018robustness}, which study the prediction mechanism of Deep Neural Networks (DNNs). In specific, for image classification, these studies find that DNNs tend to use two main kinds of features in images for prediction: (1) features which are obviously comprehensible by human eyes, such as the  presence of ``a tail'' or ``ears'' in the images of ``cat''; as well as (2) features which are incomprehensible to human but also very helpful for DNNs prediction. Both types of features make a great contribution for DNN models to make correct predictions. While, in our studied problem to protect private information, it is evident that the private features in images only reside in the features that are comprehensible by human eyes. For example, the face recognition tasks~\cite{jain2011handbook,zhao2003face}, the facial characteristics of the clients can be easily comprehended by human eyes. Therefore, we propose a defense method, Learning to Obscure Data (LODA), which optimizes to learn images that only contain the nonprivate (incomprehensible) features in the original private images of clients. Then, these learned images can serve as the clients' encrypted training samples in federated learning. From our extensive experiments, we verify that LODA presents a good protection of privacy compared to baseline methods, while LODA can preserve a high accuracy. 
The contributions of our proposed method can be summarized as:

\begin{itemize}[leftmargin=0.3in]
    \item We propose a novel privacy preserving framework, LODA, which optimizes to eliminate the private information while maintaining the nonpriviate predictive information in clients' data. 
    \item From the experimental results, our proposed defense method LODA presents high accuracy as well as reliable privacy preserving, compared to baseline methods. We also devise an adaptive attack method, aiming to defeat our defense method LODA. Similarly, LODA also demonstrates high performance against the adaptive attack, which further ensures the reliability of LODA.
    \item The proposed method LODA does not involve an expensive computational overhead. It only requires $\sim 500$ steps of gradient descent for each client image once for all without further interruption in federated learning. Thus, it is feasible to implement in practice.
\end{itemize}

\section{Related Work}\label{sec:related}
In this section, we review related works in federated learning, gradient leakage attacks and defenses. 

\textbf{Federated Learning. }
Federated learning ~\cite{shokri2015privacy,mcmahan2017communication, bonawitz2019towards, kairouz2019advances} has recently become a popular machine learning framework, which collaboratively trains machine learning models over multiple clients without uploading their data to the central server. Specifically, the clients download the global model and use the private data to calculate the local gradients or model update. Then, the local gradients or model updates are uploaded to the server and aggregated to update the global model parameters. In this process, only clients' gradient information or model update will be shared with the central server. Therefore, the clients' data is believed to be protected and Federated learning has been considered as the emerging solution for privacy protection.

\textbf{Gradient Leakage Attacks.} Federated learning may still fail to fully protect the user's privacy. 
Recently proposed gradient leakage attack methods ~\cite{zhu2020deep, geiping2020inverting, yin2021see, balunovic2021bayesian} have shown that a malicious attacker is capable of reconstructing clients' local data by exploiting the shared gradients or model updates. For example, the work~\cite{zhu2020deep} searches input data samples which have gradients with minimal Euclidean distance to the true gradients shared by clients. 
Since the method~\citep{zhu2020deep} utilizes L-BFGS~\citep{liu1989limited} as the optimization solver, which requires the smoothness of the network, their attack can not perform well in deep models and nonsmooth models. 
The attack method~\cite{geiping2020inverting} solves a similar optimization problem to reconstruct clients' data by maximizing the cosine-similarity between the gradients of the generated samples and ground truth gradients. 
Another work~\cite{yin2021see} improves the previous attacks on high-resolution images, by leveraging the matching of the batch normalization statistics between generated samples and ground truth, and also stresses the group consistency among different candidates.  A more recent work ~\cite{balunovic2021bayesian} ensembles existing attacks~\cite{zhu2020deep, geiping2020inverting} to break those defenses that are based on Gaussian noise perturbation on gradient via a Bayesian framework and gives Bayes optimal attacks for different defenses. 

Another line of works consider the scenario that the attacker is interested in exploiting the label information of the clients' data in federated learning. For example, the method~\cite{zhao2020idlg} proposes to extract label information based on the relation among different rows of the gradient matrix of the last linear layer. However, the relation is only satisfied for a batch of single image. The work~\cite{yin2021see} extends to the case of a batch of multiple images, but the argument requires strong assumptions that the batch does not contain repeated labels. In this paper, our discussion is mainly about the input sample reconstruction attacks, and we leave the label inference attack for future study.

\textbf{Defenses against Gradient Leakage Attacks.}
\label{sec:prior_defense}
To protect privacy against gradient leakage attacks in federated learning, 
various types of defenses have been proposed. They can be briefly categorized into three categories: \textbf{(1) gradient perturbation} and \textbf{(2) input encryption} and \textbf{(3) gradient encryption}. In this paper, we only discuss the first two categories of defenses, as the gradient encryption defenses usually require special setups and can be costly to implement in practice.

\textit{Gradient perturbation} aims to manipulate the shared gradients in federated learning, in order to defeat the gradient leakage attacks. For example, Gradient Pruning~\citep{zhu2020deep} prunes the small values in the shared gradients and demonstrates that pruning $70\%$ of the gradients can sufficiently destroy the data reconstruction by attackers. However, a subsequent work~\cite{huang2021evaluating} shows that gradient pruning needs to prune at least $99.9\%$ of gradients to resist stronger attacks~\cite{geiping2020inverting}, which will cause a great degradation of the trained model's accuracy. Another work called Soteria~\cite{sun2021soteria} considers to only prune a single layer of the gradient. However, it can also be circumvented by the attacking strategy which drops the matching of the pruned layer~\cite{balunovic2021bayesian}. DP-SGD ~\cite{abadi2016deep}, which clips the gradient norm and adds Gaussian noise to the gradients, provides strong theoretical guarantee for differential privacy but also results in poor accuracy especially for high-dimension data.

\textit{Input encryption} aims to encrypt and encode the clients' data so that they don't contain semantic information in the original samples. For instance, MixUp~\citep{zhang2017mixup} creates encrypted samples via linearly combining sample pairs. Motivated by Mixup, InstaHide~\cite{huang2020instahide} linearly combines private labeled images and public unlabeled images, and then randomly flips half of the sign of the pixel of the combined images. However, the adaptive attack~\cite{carlini2020attack} can successfully break InstaHide~\citep{huang2020instahide}, by training a decoding network and solving a linear regression problem. 

In this paper, our proposed defense method falls into the category of input encryption defense. However, different from existing defense methods, we encode the privacy-preserved samples by solving an innovative optimization problems that aims to hide the data but maintain useful hidden features in order to maximally guarantee the model performance while preserving privacy simultaneously.

\section{Threat Model}\label{sec:threat}

\begin{figure*}[t]
  \centering
  \includegraphics[width=0.8\linewidth]{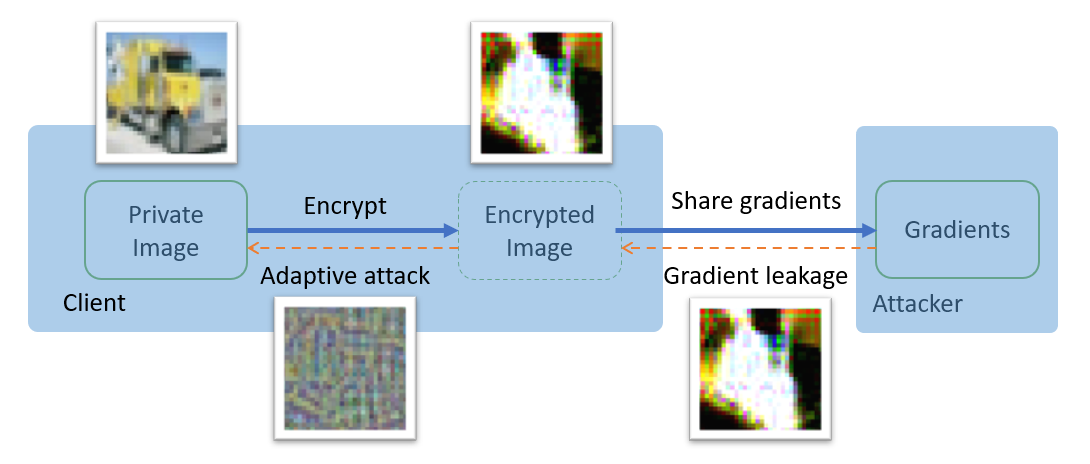}
  \caption{Gradient leakage attack and adaptive attack in federated learning. Blue arrows represent where defenses can be added (e.g., encoding or gradient pruning). Red arrows represent the gradient leakage attack and adaptive attack. The attached images show the results of each step. 
  }
  \label{fig:sc_img}
  \vspace{-0.1in}
\end{figure*}

In this section, we provide a detailed introduction about the threat model we consider in this paper. A conceptual diagram of the threat model is given in Figure ~\ref{fig:sc_img}. Before we discuss the details, we first introduce some necessary notations used in this paper. We consider classification problems for image data $(x,y) \in \mathcal{X}\times\mathcal{Y}$, where $\mathcal{X}$ and $\mathcal{Y}$ are the space of input images and labels. Here, we assume $\mathcal{X}\subseteq \mathbb{R}^d$, where $d$ is the dimension of the input data.
We denote the classifier built on a neural network as $f(\cdot)$ with parameter $\theta$, and use $\mathcal{L}(\cdot,\cdot)$ as the loss function, such as cross entropy loss.

\subsection{Gradient Leakage Attacks}

In gradient leakage attacks~\cite{zhu2020deep, geiping2020inverting}, we assume that the attacker has access to the gradient information  $\nabla_{\theta} \mathcal{L}(f(x^*;\theta),y^*)$ of a batch of private data $(x^*, y^*) \in \mathbb{R}^{b\times d} \times \mathbb{R}^b$ which is shared to the central server in federated learning. $b$ is the batch size and $d$ is the dimension of input data. The attacker is interested in reconstructing the input data $x^* \in \mathbb{R}^{b \times d}$ in order to steal data privacy. In order to achieve this goal, gradient leakage attacks attempt to search for data $x \in \mathbb{R}^{b \times d}$ with similar gradients to that of $(x^*, y^*)$.
Specifically, the gradient leakage attack can be formulated as an optimization problem that reconstructs the data $x \in \mathbb{R}^{b \times d}$: 
\begin{align}\label{eq:gla}
    \argmin_{x} D\Big(\nabla_{\theta} \mathcal{L}(f(x;\theta),y), \nabla_{\theta} \mathcal{L}(f(x^*;\theta),y^*)\Big) + \alpha \cdot \mathcal{R}(x)
\end{align}
where $D(\cdot,\cdot)$ is a distance metric that enforces $x$ to have similar gradients with $x^*$. $\mathcal{R}(\cdot)$ regularizes the reconstructed data based on the data prior information, and $\alpha$ controls the trade-off between reconstruction and regularization. In Eq.~\eqref{eq:gla}, the label $y$ of the reconstructed data $x$ can be assumed to be known by the attacker~\citep{geiping2020inverting}, or it can be inferred by label inference attack~\cite{zhao2020idlg} with close to perfect accuracy. In our paper, we also assume $y = y^*$ following previous works~\citep{geiping2020inverting, zhang2017mixup}. Various attack methods have different choices of the distance metric and regularization. The state-of-the-art attacks~\citep{geiping2020inverting} define $D(\cdot,\cdot)$ as the negative cosine similarity and $\mathcal{R}(\cdot)$ as the total variation of images.

\subsection{Adaptive Attack on Input Encryption}

We also introduce one representative defense method,  InstaHide~\cite{huang2020instahide}, which is devised to resist gradient leakage attacks. It is a state-of-the-art input encryption defense method (see Section~\ref{sec:prior_defense}). 

\textbf{InstaHide}~\cite{huang2020instahide} aims to encrypt one private sample by linearly combining it with other samples in the private dataset or a large public dataset. In detail, consider the scenario that given a batch of private images $\{x_k\}_{k=1}^{n}$. For each private image $x^*$ in the batch, the encrypted image $\mathcal{E}(x^*)$ is obtained by linearly combing $x^*$ and other samples, following the strategy:
\begin{align}\label{eq:instahide}
\mathcal{E}(x^*; x_1, \dots, x_n) = \sigma( \sum_{k=1}^n \alpha_k x_k),    
\end{align}
where $\alpha_1,\dots,\alpha_n$ are nonnegative coefficients that sum to 1. The function $\sigma(\cdot)$ is an element-wise operator that randomly flips the sign of each pixel value with a $50$\% chance. InstaHide~\citep{huang2020instahide},  also encrypts the label $y^*$ using the same linear combination (with the same coefficients $\alpha_1, \dots,\alpha_n)$ corresponding to the encrypted input samples. If we denote the private batch as $X = [x_1, x_2, \dots, x_n]^T \in \mathbb{R}^{n \times d}$ and $M = [\bm{\alpha_1}, \bm{\alpha_2}, \dots, \bm{\alpha_n}]^T \in \mathbb{R}^{n\times n}$, then we can reformulate the strategy in Eq.~\eqref{eq:instahide} by matrix form: 
$$\mathcal{E}(X) = \sigma(M X).$$

\textbf{Adaptive Attack to Break InstaHide.}
In~\citep{carlini2020attack}, a strategy is proposed to break Instahide and successfully de-encrypt the samples generated by InstaHide. Briefly speaking, the method~\citep{carlini2020attack} first trains a neural network to predict whether two encrypted images $\mathcal{E}(x_1)$ and  $\mathcal{E}(x_2)$ contain the same private image, so it can roughly estimate the matrix $M$ and get $\hat{M}$ (more details can be found in~\cite{carlini2020attack}). Then, the attack repeatedly solves the regression problem: 
\begin{align}\label{eq:carliniattack}
    \argmin_{X'} ||abs\big(\hat{M} X'\big) - abs\big(\mathcal{E}(X)\big)||^2
\end{align}
until it can get reconstructed samples $X'$. In Eq.~\eqref{eq:carliniattack}, the attack~\citep{carlini2020attack} also introduces an extra operator $abs(\cdot)$ to get the absolute value for each pixel, to nullify the random sign flipping operator $\sigma$ in InstaHide. Intuitively, since InstaHide~\cite{huang2020instahide} linearly combines the private samples $x^*, x_1, ..., x_n$, the attack~\cite{carlini2020attack} aims to reverse the combination process by decomposing the encrypted samples via solving regression problems as in Eq.~\eqref{eq:carliniattack}.
From the results in~\citep{carlini2020attack}, the attack can successfully reconstruct the original samples when the attacker has access to the encrypted images.

\section{Defense by Learning to Obscure Data}\label{sec:defend}

As discussed in Section~\ref{sec:related}, InstaHide ~\cite{huang2020instahide} cannot adequately protect the privacy of images because it encrypts private samples only via linearly combining them with other private (or public) samples. Thus, InstaHide is still vulnerable to the adaptive attack~\cite{carlini2020attack} which reverses the combination process and decomposes the encrypted samples. In this section, we propose a novel input encryption method, called \textbf{\textit{Learning to Obscure Data (LODA)}}. In our proposed defense, we encrypt each sample by adding an sophiscatedly optimized perturbation that is generated by learning to obscure the private information in the original images. Next, we will introduce our proposed method LODA in detail.

\subsection{Learning to Obscure Data (LODA)}

Our method is motivated by previous works~\cite{ilyas2019adversarial, tsipras2018robustness, shafahi2018poison} which study the prediction mechanism of Deep Neural Networks (DNNs). Specifically, these works suggest that DNNs tend to use two main kinds of features in images for prediction: \textbf{(1)} features which are \textbf{obviously comprehensible} by human eyes, such as the  presence of ``a tail'' or ``ears'' in the images of ``cat''; and \textbf{(2)} features which are \textbf{ incomprehensible} to human eyes. These features contain signals that are not visually relevant to the label ``cat'' or even visually imperceptible by human eyes. In fact, from the works~\cite{ilyas2019adversarial, tsipras2018robustness, shafahi2018poison}, they find the fact that a DNN model which only depends on incomprehensible features is suffice to achieve a high prediction accuracy.
This fact motivates us to design an privacy protection method that disentangles these two types of features such that the privacy can be protected while keeping useful features to train the model that maintains a high prediction accuracy. In our work, we assume that the private information of clients, e.g., the facial characteristics in face recognition tasks, only resides in comprehensible features as defined in~\cite{ilyas2019adversarial, tsipras2018robustness, shafahi2018poison}.
We give a definition for ``private feature'' as:

\begin{definition}\textbf{(Private Feature)}
\textit{A feature which is visually comprehensible by human eyes and contains private information of a client.}
\end{definition}

Then, we desire to encrypt the private image to get a privacy-preserved image, by excluding its private features and only keeping non-private (incomprehensible) but predictive features:
\begin{definition}\textbf{(Non-private Predictive Feature)}
\textit{A feature which is visually incomprehensible by human eyes but is helpful for model prediction.}
\end{definition}

Therefore, for federated learning which trains on clients' data solely with non-private (incomprehensible) predictive features, Gradient Leakage Attacks cannot obtain (comprehensible) private information of the clients. In our work, to obtain such images, we cannot directly manipulate the pixels of very complex, high-dimensional
datasets. Instead, we will leverage an auxiliary neural network $g(\cdot)$ which is pretrained on a public dataset. Then, we propose the following optimization problem to encrypt each private image $x^*$ and get the encrypted image $\mathcal{E}(x^*)$:
\begin{align}\label{eq:loda}
    \mathcal{E}(x^*) = \argmin_{x'\in \mathcal{X}} ||g(x')-g(x^*)||^2 - c\cdot ||x'-x^*||^2
\end{align}
where $x^*$ is the original private image, $\mathcal{X}$ is the space of valid images, $g(\cdot)$ is the hidden layer activation values of the pretrained model, and $c$ is a positive hyperparameter. In particular, in order to maintain predictive features, the first term in Eq.~\eqref{eq:loda} enforces that the encrypted image $x'$ has maximal feature alignment with the original image $x^*$. In order to protect data privacy, the second term enforces the encrypted image to be significantly different from private image $x^*$. In this work, we adopt gradient decent to solve Eq.~\eqref{eq:loda} and the detailed algorithm is shown in Appendix~\ref{app:LODA_alg}. Next, we briefly discuss why solving Eq.~\ref{eq:loda} can effectively exclude private features while maximally including non-private predictive features.

\textbf{Private Feature Elimination.} 
To solve the optimization problem in Eq.~\eqref{eq:loda}, we initialize the starting point of $x'$ from an image randomly sampled from a public dataset.
This will ensure the initialization of $x'$ does not contain any private (comprehensible) feature of $x^*$. Moreover,
in Eq.~\eqref{eq:loda},
the second term $-c ||x' - x^*||_2$ maximizes the Euclidean distance between the perturbed image $x'$ and the original private sample $x^*$. This will ensure the image $x'$ to be not visually similar to $x^*$ through the whole optimization process.
As a result, both the Euclidean penalty and initialization of $x'$ will enforce the generated sample $x'$ to be far way from the private image $x^*$ such that it eliminates the information in the original image $x^*$ that is comprehensible by human eyes. Thus, the encrypted image $\mathcal{E}(x^*)$ does not contain private features in $x^*$.

\textbf{Non-private Predictive Feature Alignment.} In Eq.~\eqref{eq:loda}, by minimizing the first term $||g(x') - g(x^*)||_2$, we enforce the perturbed image $x'$ to maintain similar hidden features as the original sample $x^*$ under the feature extractor of the pretrained model. In particular, we specifically define $g(\cdot)$ as the
output values of all (or part of) convolutional layers
in a CNN based model, such as ResNet18 \citep{he2016deep}. This process can help the perturbed sample $x'$ maintain hidden features in $x^*$
such that it preserves the predictive power. Remind that the second term in Eq.~\eqref{eq:loda} and the initialization of $x'$ act to exclude private features in the input pixel space. Thus, all features other than private features, such as non-private predictive features will be included in $x'$ in the final encrypted image $\mathcal{E}(x^*)$.

\textbf{Remark.} In practice, we encrypt each image in the private dataset using the optimization method proposed in Eq.~\eqref{eq:loda}. Then, the clients and server can conduct federated learning on our encrypted dataset and enjoy good accuracy and privacy. Compare to the previous defense methods, such as InstaHide~\cite{huang2020instahide}, our method directly perturbs the features of each private image, instead of stacking the private image with other images. Therefore,  
our defense can be hardly defeated by attacks which tries to decompose the encrypted images such as~\citep{carlini2020attack}. Compare to gradient perturbation methods such as~\cite{zhu2020deep}, which naively trim the shared gradient information, our optimization process can also guarantee the model accuracy while demanding privacy protection.

\subsection{Adaptive Attack against LODA}

Although promising, it is still unknown whether LODA can be defeated by adaptive attacks that are devised specifically to attack LODA. In this subsection, we design an adaptive attack against LODA.

We consider the scenario that the attacker tries to reconstruct the original images $x^*$ by searching the root $x^*$ for Eq.~\ref{eq:loda}.
Suppose that given the access to the feature extractor $g(\cdot)$ and public data used in the initialization, the adaptive attack for LODA aims to recover the image $x^*$ from encrypted image $x'=\mathcal{E}(x^*)$. The obscured data is obtained by solving Eq.~\eqref{eq:loda}, which is a minimization problem on $x'$. 
Thus, if the obscured data $x'$ reaches the optimal point, it satisfies the stationary condition:  
\begin{align} \label{eq:adaptive1}
    \nabla g(x'))^T \Big(g(x') - g(x^*)\Big) - c(x'-x^*) = 0.
\end{align}
In Eq.~\eqref{eq:adaptive1}, $\nabla g(x')$ is the gradient of feature extractor $g(\cdot)$ evaluated at the encrypted image $\tilde x$, which is accessible by the attacker. Thus, $x^*$ is the root of Eq.~\eqref{eq:adaptive1}. 
Therefore, to recover the private image $x^*$, we design an adaptive attack that solves the following problem:
\begin{align} \label{eq:adaptive1_min}
    {x}^* = \argmin_{x\in \mathcal{X}} \|\nabla g(x')^T \Big(g(x') - g(x)\Big) - c(x'-x)\|_2^2
\end{align}
where the encrypted image $x'$ is given and the minimization is with respect to the private image $x$. $c$ can be regarded as a hyperparameter to search for the best adaptive attack. Eq.~\eqref{eq:adaptive1_min} can be solved by gradient descent. In Section~\ref{sec:experiment}, we conduct experiments to verify that LODA also achieves good privacy preserving even when it is attacked by this adaptive attack strategy.

\section{Experiment}\label{sec:experiment}

In this section, we conduct experimental studies to verify the effectiveness of our proposed framework LODA. In detail, we will first introduce the basic experimental setup. Then, we compare LODA with baseline methods, in terms of model accuracy and privacy protection when defending against Gradient Leakage Attacks. Moreover, we will also evaluate their privacy protection when they are under the threats of adaptive attacks.

\subsection{Experimental Setup}

The experiment evaluates the protection of different defense methods against gradient leakage attack on CIFAR-10 \citep{krizhevsky2009learning} and SHVN \citep{37648} datasets, whose training sets are assumed to be private. 

\textbf{Federated Learning Setup.} In federated learning, the potential leakage happens when the clients exchange gradients calculated based on their private data. In our experiment, we consider the threat model as that given the gradient calculated from a private image, the attacker performs gradient leakage attack to reconstruct the image, i.e., the case for batch size $= 1$. 

\textbf{Defenses and Attacks.} We consider the following defenses and corresponding attacks:
\begin{itemize}[leftmargin=0.3in]
\item \textit{No Defense.} the original federated learning without applying any defense methods.
    \item \textit{Gradient Pruning.} Gradient pruning sets the smallest $p$ percent of the gradient elements to $0$. In this case, the gradient leakage attack is adapted to match only the unpruned part.
    \item \textit{Mixup.} Mixup linearly combines the private image with $k-1$ other images from the batch. The attacker first recovers the encrypted images from the gradients. In addition, with an adaptive attack, the attacker further decodes the private images from the encrypted images. Without adaptive attack, the privacy evaluation is based on the comparison between encrypted image and one of its Mixup components with the largest weight. We include the cases $k = \{4, 6\}$.
    \item \textit{InstaHide.} InstaHide applies a random sign flip operation on top of Mixup. The evaluation is similar to the mixup setting. The adaptive attack follows~\cite{carlini2020attack}. We include the cases $k = \{4, 6\}$.
    \item \textit{LODA.} LODA learns the nonprivate predictive features through feature alignment, meanwhile discarding private features through pixel distance maximization. Adaptive attack is based on the formulation Eq. ~\ref{eq:adaptive1_min}. Without adaptive attack, the privacy evaluation is based on the comparison between the recovered images from the gradients and private images. Next we introduces the setting for different variants of LODA.
\end{itemize}

\textbf{Setup of LODA.} LODA varies with the use of different $g(\cdot)$ in the feature alignment term of Eq.~\ref{eq:loda}. For CIFAR-10, we use a ResNet18 trained on CIFAR-100 (assume public) as the feature extractor. We consider two cases that balance the trade-off between privacy and accuracy: (1) $\text{LODA}_1$: $g(\cdot)$ is the output of first five convolutional layers of the feature extractor and $c=20$; (2) $\text{LODA}_2$: $g(\cdot)$ includes the output of all the convolutional layers of the feature extractor and $c=30$. The choice of $c$ is discussed in Appendix ~\ref{app:ablation}. The initialization in Algorithm ~\ref{algo:loda} is an image randomly sampled from the training set of CIFAR-100. We use $500$ iterations for the normalized gradient descent with learning rate $0.1$ in the generation of encrypted image. 
    
For SHVN, we use a ResNet18 trained on sub-sampled extra dataset (assume public) with the same size as the training dataset as the feature extractor. We denote as $\text{LODA}_3$: $g(\cdot)$ is the output of last five convolutional layers of the feature extractor $+$ linear layers and $c=1$. The initialization is an image sampled from the training set of CIFAR100.

\textbf{Evaluation Metrics.} The evaluation involves two aspects: privacy and accuracy. For privacy, we use signal-to-noise ratio (PSNR) and learned perceptual image patch similarity (LPIPS)~\cite{zhang2018perceptual} to measure the mismatch between reconstructed images and private images. If the reconstructed images are visually dissimilar from private images, PSNR will be smaller and LPIPS will be larger, 
i.e., the most severe privacy leakage happens at the Max PSNR and Min LPIPS.

\textbf{Gradient Leakage Attack.} 
In the evaluation, the attack method follows ~\cite{geiping2020inverting}, using cosine similarity and total variation regularization. The weight of total variation $\alpha_{TV}$ are searched over the space $\{0.00001, 0.0001, 0.001, 0.01, 0.1\}$. For CIFAR10, We use $\alpha_{TV} = 0.0001$ for no gradient pruning and $\alpha_{TV} = 0.001$ for defenses with gradient pruning. For SVHN, we use $\alpha_{TV} = 0.01$ for all cases.

\textbf{Adaptive Attacks.} The adaptive attack for Mixup/InstaHide follows ~\cite{carlini2020attack} where we combine the reconstructed images from $50$ iterations. We use Adam with learning rate $0.01$ to solve the regression problem Eq. ~\ref{eq:carliniattack}. The adaptive attack for LODA was introduced in Section 4.2. We use the same feature extractor as that used in the image encryption. The pixel weight $c$ is regarded as a hyperparameter. It is searched over the space $\{0, 1, 5, 10, 20\}$ and is set to $1$ in the following experiments. For the step number of gradient descents used in the adaptive attack, we search over $\{50, 100, 200, 400\}$ and set it to $100$ in the following experiments.

\subsection{Accuracy and Privacy Comparison}

In Table~\ref{tab:cifar10no}, we present the experimental results for different defenses under gradient leakage attack. In Table~\ref{tab:cifar10after}, we present the experimental results for different defenses under gradient leakage attack and the corresponding adaptive attacks. From these two tables we can make the following observations:
\begin{itemize}[leftmargin=0.3in]
    \item The privacy protection of gradient pruning is very bad when $p=0.9$. When $p=0.99$, it hides more gradient elements such that the privacy protection becomes better but the price is to significantly sacrifice the accuracy.
    
    \item Mixup protects privacy well under gradient leakage attack while achieving reasonable accuracy. However, Table~\ref{tab:cifar10after} shows that Mixup loses almost all privacy under stronger adaptive attack~\cite{carlini2020attack}.
    
    \item InstaHide applies a random sign flip on top of Mixup. Therefore, it protects privacy much better than Mixup and gradient pruning but it has lower accuracy because of the random sign flipping. With stronger adaptive attack~\cite{carlini2020attack}, the privacy protection of InstaHide becomes much worse, which clearly demonstrates its vulnerability.
    
    \item The proposed LODA achieves comparable privacy protection and accuracy as InstaHide under gradient leakage attack. However, LODA protects privacy much better than InstaHide under stronger adaptive attack.
\end{itemize}

\begin{table*}[h] 
  \caption{Accuracy and privacy under gradient leakage attack on CIFAR10.}
  \label{tab:cifar10no}
  \resizebox{\columnwidth}{!}{
  \begin{tabular}{c|c|cc|cc|cc|cc }
    \toprule
    \textbf{Method} &\textbf{No Defense}&
    \multicolumn{2}{c|}{\textbf{GradPruning}} & 
    \multicolumn{2}{c|}{\textbf{Mixup}}&
    \multicolumn{2}{c|}{\textbf{InstaHide}} &
    \multicolumn{2}{c}{\textbf{LODA}}\\\hline\hline
    \textbf{Parameter} & - & p=0.9  & p=0.99 & k=4 & k=6 & k=4 & k=6 & $LODA_1$  & $LODA_2$ \\\hline
    \textbf{Accuracy (\%)} & \textbf{95.20} & 93.30 & 88.64 & 93.84 & 92.38 & 91.80 & 88.52 & 89.50 & 91.78 \\\hline
    \textbf{Avg. LPIPS ($\uparrow$)} & 0.098 $\pm$ 0.062 & 0.224$\pm$0.097  & 0.490$\pm$0.087 & 0.377 $\pm$ 0.119 & 0.400 $\pm$ 0.079 & \textbf{0.656 $\pm$ 0.029} & 0.628 $\pm$ 0.042 & 0.589$\pm$0.050 & 0.566 $\pm$ 0.057\\
    \textbf{Min LPIPS} & 0.016 & 0.074 & 0.300 & 0.091 & 0.224 & \textbf{0.607} & 0.560 & 0.480 &0.423\\\hline
    \textbf{Avg. PSNR ($\downarrow$)}& 16.61 $\pm$2.62 &11.48 $\pm$2.36 & \textbf{5.44 $\pm$ 2.06} & 15.01$\pm$2.02 & 14.2$\pm$2.25 & 9.51$\pm$1.36 & 9.98$\pm$1.32 & 5.77$\pm$1.46 & 7.41 $\pm$ 2.49  \\
    \textbf{Max PSNR} & 20.78 & 15.52 & 10.43 & 18.02 & 18.03 & 12.87 & 12.25 & \textbf{7.77} & 12.08\\
    \hline
\end{tabular}
}
\vspace{-0.1in}
\end{table*}

\begin{table*}[h]
  \caption{Accuracy and privacy under gradient leakage attack + adaptive attack on CIFAR10.}
  \label{tab:cifar10after}
  \resizebox{\columnwidth}{!}{
  \begin{tabular}{c|c|cc|cc|cc|cc }
    \toprule
    \textbf{Method} &\textbf{No Defense}&
    \multicolumn{2}{c|}{\textbf{GradPruning}} & 
    \multicolumn{2}{c|}{\textbf{Mixup}}&
    \multicolumn{2}{c|}{\textbf{InstaHide}} &
    \multicolumn{2}{c}{\textbf{LODA}}\\\hline\hline
    \textbf{Parameter} & - & p=0.9  & p=0.99 & k=4 & k=6 & k=4 & k=6 & $LODA_1$  & $LODA_2$ \\\hline
    \textbf{Accuracy (\%)} & \textbf{95.20} & 93.30 & 88.64 & 93.84 & 92.38 & 91.80 & 88.52 & 89.50 & 91.78 \\\hline
    \textbf{Avg. LPIPS ($\uparrow$)} & 0.098 $\pm$ 0.062 & 0.224$\pm$0.097  & 0.490$\pm$0.087 & 0.039 $\pm$ 0.028  & 0.049 $\pm$ 0.030 & 0.395$\pm$0.075 & 0.394$\pm$ 0.075 & 0.537 $\pm$ 0.058  &  \textbf{0.558 $\pm$ 0.057} \\
    \textbf{Min LPIPS} & 0.016 & 0.074 & 0.300 & 0.016 & 0.015 & 0.273& 0.233& 0.452 & \textbf{0.456} \\\hline
    \textbf{Avg. PSNR ($\downarrow$)}& 16.61 
    $\pm$2.62 &11.48 $\pm$2.36 & 5.44 $\pm$ 2.06 & 29.23 $\pm$ 2.24 & 28.25 $\pm$ 2.26 &10.77$\pm$ 3.43 & \textbf{10.62 $\pm$ 3.12} & 12.95 $\pm$ 1.66 & 13.04 $\pm$ 2.38  \\
    \textbf{Max PSNR} & 20.78 & 15.52 & \textbf{10.43} & 34.04 & 31.94 & 18.72&18.86 & 16.24 & 16.91\\
    \hline
\end{tabular}
}
\vspace{-0.1in}
\end{table*}

\begin{table*}[h] 
  \caption{Accuracy and privacy under gradient leakage attack on SVHN.}
  \label{tab:svhnno}
  \resizebox{\columnwidth}{!}{
  \begin{tabular}{c|c|cc|cc|cc|c }    
  \toprule
    \textbf{Method} &\textbf{No Defense}&
    \multicolumn{2}{c|}{\textbf{GradPruning}} & 
    \multicolumn{2}{c|}{\textbf{Mixup}}&
    \multicolumn{2}{c|}{\textbf{InstaHide}} &
    \multicolumn{1}{c}{\textbf{LODA}}\\\hline\hline
    \textbf{Parameter} & - & p=0.9  & p=0.99 & k=4 & k=6 & k=4 & k=6 & $LODA_3$  \\\hline
    \textbf{Accuracy (\%)} & \textbf{96.33} & 95.98 & 95.24 & 94.99 & 94.02 & 94.93 & 92.42 & 95.29 \\\hline
    \textbf{Avg. LPIPS ($\uparrow$)} & 0.168 $\pm$ 0.088 & 0.287 $\pm$ 0.102  & 0.495 $\pm$ 0.067 &  0.341 $\pm$ 0.109 & 0.340 $\pm$ 0.115 & 0.694 $\pm$ 0.036 & 0.696 $\pm$ 0.036  &  \textbf{0.715 $\pm$ 0.049}  \\
    \textbf{Min LPIPS} & 0.034 & 0.073 & 0.361 & 0.135 & 0.085 & 0.622 & 0.603 & \textbf{0.648}\\\hline
    \textbf{Avg. PSNR ($\downarrow$)}&  23.44 $\pm$ 3.14 & 20.51 $\pm$ 3.00 & 16.08 $\pm$ 2.53 & 16.37 $\pm$ 2.49 & 15.79 $\pm$ 2.33 & 10.43 $\pm$ 0.81 & 10.46 $\pm$ 1.08 & \textbf{8.46 $\pm$ 1.48}\\
    \textbf{Max PSNR} & 30.63 & 26.29 & 20.88 & 21.54 & 21.59 & 12.05 & 12.85 & \textbf{11.52} \\
    \hline 
\end{tabular}
}
\vspace{-0.1in}
\end{table*}

\begin{table*}[ht!] 
  \caption{Accuracy and privacy under gradient leakage attack + adaptive attack on SVHN.} 
  \label{tab:svhnafter}
  \resizebox{\columnwidth}{!}{
  \begin{tabular}{c|c|cc|cc|cc|c }
    \toprule
    \textbf{Method} &\textbf{No Defense}&
    \multicolumn{2}{c|}{\textbf{GradPruning}} & 
    \multicolumn{2}{c|}{\textbf{Mixup}}&
    \multicolumn{2}{c|}{\textbf{InstaHide}} &
    \multicolumn{1}{c}{\textbf{LODA}}\\\hline\hline
    \textbf{Parameter} & - & p=0.9  & p=0.99 & k=4 & k=6 & k=4 & k=6 & $LODA_3$  \\\hline
    \textbf{Accuracy (\%)} & \textbf{96.33} & 95.98 & 95.24 & 94.99 & 94.02 & 94.93 & 92.42 & 95.29 \\\hline
    \textbf{Avg. LPIPS ($\uparrow$)} & 0.168 $\pm$ 0.088 & 0.287 $\pm$ 0.102  & 0.495 $\pm$ 0.067 & 0.046 $\pm$ 0.024  & 0.058 $\pm$ 0.026 & 0.286 $\pm$ 0.197 & 0.243 $\pm$ 0.169 & \textbf{0.6041 $\pm$ 0.0468}  \\
    \textbf{Min LPIPS} & 0.034 & 0.073 & 0.361 & 0.016 & 0.016 & 0.031 & 0.067 & \textbf{0.5191}\\\hline
    \textbf{Avg. PSNR ($\downarrow$)}&  23.44 $\pm$ 3.14 & 20.51 $\pm$ 3.00 & 16.08 $\pm$ 2.53 & 24.71 $\pm$ 2.65 & 24.44 $\pm$ 3.07 & 16.96 $\pm$ 4.90 & 18.95 $\pm$ 4.24 & \textbf{13.35 $\pm$ 2.52} \\
    \textbf{Max PSNR} & 30.63 & 26.29 & 20.88 & 28.42 & 29.13 & 24.80 & 23.62 & \textbf{19.02}\\
    \hline 
\end{tabular}
}
\end{table*}

\vskip0.1in

\begin{figure*}[ht!]
  \centering
  \includegraphics[width=1\linewidth]{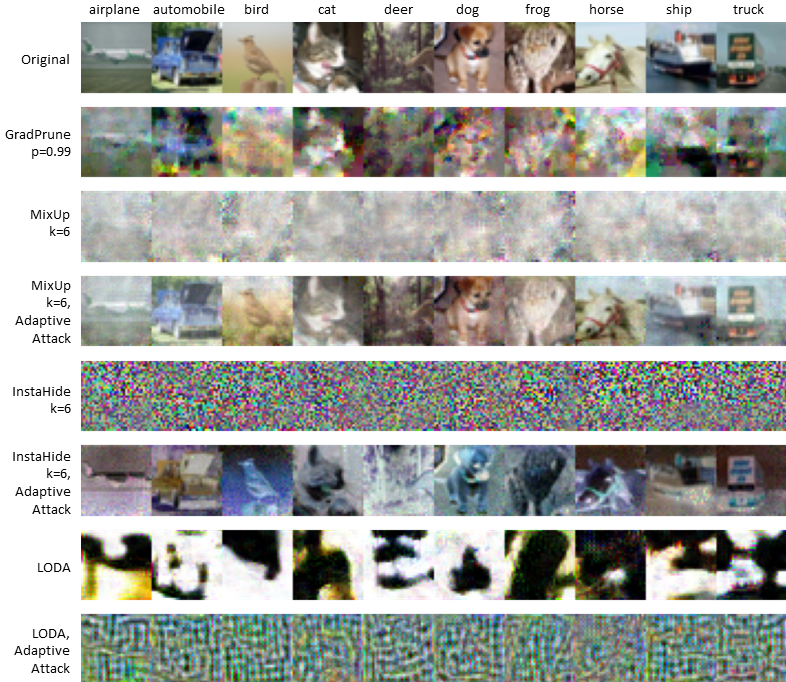}
  \caption{Privacy attack results for each class of CIFAR10 for  different defenses. The first row shows the private images. The second to fifth rows correspond to GradPruning ($p=0.99$), Mixup ($k=6$), InstaHide ($k=6$) and $LODA_1$, respectively. }
  \label{fig:all_images}
    \vskip-0.2in
\end{figure*}

We also present the experimental results for SVHN in Table~\ref{tab:svhnno} and Table~\ref{tab:svhnafter} under the gradient leakage attack and gradient leakage attack + adaptive attack, respectively. Although the adaptive attack for $\text{LODA}_3$ weakens the privacy protection, $\text{LODA}_3$ still remains the most effective defense method.

\subsection{Encrypted Images of LODA}

In order to provide a visual comparison of the privacy protection between different defense and attack algorithms, we randomly select one image from all the classes in the CIFAR10 dataset and show the original images, protected images, and recovered images (including the gradient leakage attack and adaptive attacks) in Figure~\ref{fig:all_images}. The hyperparameters for each defense are chosen so than they have similar accuracy for the retrained model (as shown in Table~\ref{tab:cifar10no}). From Figure~\ref{fig:all_images}, we can make the following observations:
\begin{itemize}[leftmargin=0.3in]
    \item Although Mixup and InstaHide can hide the private image under gradient leakage attack, the recovered images are very close to the private images under stonger adaptive attack. These suggest that Mixup and InstaHide are still vulnerable to adaptive attack and will leak privacy. 
    
    \item The proposed LODA protects the private image well under both gradient leakage attack and adaptive attack, which demonstrates its effectiveness.
    
    \item The visual perception about privacy protection qualitatively matches well with the numerical measurements such as PSNR and LPIPS in Tables~\ref{tab:cifar10no},~\ref{tab:cifar10after},~\ref{tab:svhnno}, and~\ref{tab:svhnafter}. This suggests the validity of these privacy measurements.
\end{itemize}

\section{Conclusion}
In this work, we propose a novel defense strategy against gradient leakage attacks via learning to obscure the data. We argue the possibility to find images that have similar hidden features as the clean images while having large distance in the pixel space. The proposed defense strategy enables us to find encrypted images such that the predictive features are preserved to train a well-behaved machine learning model while largely protecting the data privacy. 
Through extensive experiments, we demonstrate that the proposed defense strategy can achieve high model performance in terms of prediction accuracy while preserving the data privacy. 
Furthermore, it outperforms state-of-the-art defense methods in terms of privacy protection under the comparable model performance. This defense method can be combined with other gradient perturbation defenses to further enhance the privacy.




\medskip

{
\small
\bibliographystyle{plain}
\bibliography{references}

}

\newpage
\appendix

\section*{Societal Impact and Limitations}

The methodology proposed in this paper might have significant positive societal impact since it
reduces machine learning models’ vulnerability to gradient leakage attacks that generally exist in federated learning applications and enhances the clients' data privacy. While we are unaware of any potential negative society
impact, we point out two limitations of this work: (1) although this defense performs well under existing attacks, it does not theoretically guarantee the privacy of gradient information. (2) this paper only examines this defense alone and this defense could be combined with gradient perturbation defenses to further enhance the privacy. It will be interesting to investigate these problems in future works.
\section{Algorithm}\label{app:LODA_alg}
Here we give the detailed explanation about Algorithm ~\ref{algo:loda}.
For each private image, we first initialize $x'$ with a random sample from some public dataset (step 3). Then, we solve the optimization problem in Eq.~\eqref{eq:loda} by normalized gradient descent and project the image back to valid image space in each step (steps 5-7). All implementation codes can be found at: 
\url{https://anonymous.4open.science/r/learning_to_obscure_data-4444}

\begin{algorithm}[th!]
\caption{ \textsc{Learning to Obscure Data (LODA)}}
\label{algo:loda}
\begin{algorithmic}[1]
\STATE\textbf{Input} :  Private dataset $\{(x_i^*, y_i^*)\}_{i=1}^N$, a public dataset, a pretrained feature extractor $g(\cdot)$, maximum training iteration $T$, a constant value $c$ and learning rate $\tau$.
\FOR{i = 1:N}
\STATE
Initialize $x'_i$ by a random sample from a public dataset.
\FOR{t=1:T}
\STATE \textbf{1.}. Calculate the gradient: \\
$grad = \nabla_{x'_i} \Big( ||g(x'_i) - g(x^*_i))||^2 - c||x'_i - x^*_i||^2 \Big)$
\STATE \textbf{2.}. Update with normalized gradient descent:\\
$x'_i = x'_i - \tau\cdot \frac{grad}{||grad||}$
\STATE \textbf{3.} Project the image to $\mathcal{X}$, $x'_i = \mathbf{proj}_{x \in \mathcal{X}}  x'_i$
\ENDFOR
\ENDFOR
\STATE \textbf{Output} : obscured data $\{(x_i', y_i^*)\}_{i=1}^N$  
\end{algorithmic}
\end{algorithm}

\section{Ablation study of LODA}\label{app:ablation}
\subsection{Impact of weight \texorpdfstring{$c$}{c}}
To get a better understanding of the trade-off between privacy and accuracy for our defense LODA, we conduct ablation studies on the weight $c$ for the distance maximization. The goal of LODA is to maintain nonprivate features that can be used to train a new model and discard private features. However, the elimination of private features can conflict with the preservation of useful features. The weight $c$ controls the relative importance of the two aspects. With increasing weight $c$, LODA emphasizes more on privacy protection and hence could hurt the accuracy.
We varies the weight $c = \{10, 20, 30, 40, 50\}$ for two LODA settings ($\text{LODA}_1$ and $\text{LODA}_2$) and present the results in Table~\ref{tab:loda_1} and Tabel~\ref{tab:loda_2} for $\text{LODA}_1$ and $\text{LODA}_2$, respectively. From both tables, we can see there indeed exits such trade-off between accuracy and privacy controlled by weight $c$. Therefore, it suggests the flexibility of our proposed defense strategy in making a good balance.

\begin{table*}[!ht] 
  \caption{Weight change for $\text{LODA}_1$}
  \label{tab:loda_1}
  \resizebox{\columnwidth}{!}{
  \begin{tabular}{c|ccccc }
  \toprule
    $\text{LODA}_1$ & \textbf{c=10} & \textbf{c=20} & \textbf{c=30} & \textbf{c=40} & \textbf{c=50}
    \\\hline\hline
    \textbf{Acc} & 93.02 & 89.50 & 83.46 & 78.48 & 71.86   
    \\\hline
    \textbf{Avg. LPIPS($\uparrow$)} & 0.539 $\pm$ 0.073 & 0.589 $\pm$ 0.050 & 0.610 $\pm$ 0.050 & 0.626 $\pm$0.055 & 0.626 $\pm$ 0.050 \\
    \textbf{Min LPIPS} & 0.341 & 0.480 & 0.497 & 0.507 & 0.528
    \\\hline
    \textbf{Avg. PSNR($\downarrow$)} & 6.39 $\pm$ 1.66 & 5.77 $\pm$ 1.46 & 5.73 $\pm$ 1.38 & 5.83 $\pm$ 1.49 & 5.82$\pm$1.38  \\
    \textbf{Max PSNR} &  9.35  & 7.77 & 7.77 & 8.39 & 7.86 \\
    \hline 
\end{tabular}
}
\end{table*}

\begin{table*}[!t]
  \caption{Weight change for $\text{LODA}_2$}
  \label{tab:loda_2}
  \resizebox{\columnwidth}{!}{
  \begin{tabular}{c|ccccc }
  \toprule
    $\text{LODA}_2$ & \textbf{c=10} & \textbf{c=20} & \textbf{c=30} & \textbf{c=40} & \textbf{c=50} \\\hline\hline
    \textbf{Acc} & 93.95 & 93.49 & 91.78 & 89.40 & 87.78   
    \\\hline
    \textbf{Avg. LPIPS($\uparrow$)} & 0.453$\pm$0.079 & 0.503 $\pm$ 0.053 & 0.566 $\pm$ 0.057 & 0.581 $\pm$0.061 & 0.602 $\pm$ 0.051 \\
    \textbf{Min LPIPS} & 0.299 & 0.408 & 0.423 & 0.445 & 0.488 \\\hline
    \textbf{Avg. PSNR($\downarrow$)} & 9.72 $\pm$ 3.04 & 8.07 $\pm$ 2.88 & 7.41 $\pm$ 2.49 & 6.55 $\pm$ 2.08 & 6.05$\pm$1.75  \\
    \textbf{Max PSNR} &  15.10  & 13.31 & 12.08 & 10.43 & 9.41* \\
    \hline 
\end{tabular}
}
\end{table*}

\subsection{Generalizability of LODA}\label{app:abl_acc}

We present the generalizability of LODA in Table ~\ref{tab:abl_acc}. Specifically, we show the test accuracy when using data generated by $\text{LODA}_1$ and $\text{LODA}_2$ to train different architectures, including ResNet18~\citep{he2016deep}, ResNet34~\citep{he2016deep}, VGG16 ~\citep{simonyan2014very}, DenseNet121~\citep{huang2017densely}. We also include the test accuracy of the model trained on original CIFAR10. From the table, we have the observations: (a) different architectures can achieve comparable test performance; (b) architectures that achieve higher (lower) accuracy on original data can also achieve higher (lower) accuracy on obscured data, e.g., VGG16 has lowest accuracy for all cases.
Thus, we conclude that the obscured data generated from LODA show good generalization ability across different architectures.

\begin{table}[th!] 
    \centering
  \resizebox{0.7\columnwidth}{!}{
  \begin{tabular}{ccccc}
  \toprule
    Dataset & \textbf{ResNet18} & \textbf{ResNet34} & \textbf{VGG16} & \textbf{DenseNet121} 
    \\\hline\hline
    \textbf{CIFAR10} & \underline{95.20} & \textbf{95.47} & 93.44 & 94.46  
    \\\hline
    $\textbf{LODA}_1$ & \textbf{89.50} & \underline{89.05} & 83.77 & 87.83  
    \\\hline
    $\textbf{LODA}_2$ & \textbf{91.78} & \underline{91.42} & 88.81 & 90.32 
    \\\hline
\end{tabular}
}
\vspace{0.1in}
\caption{Cross-architecture test accuracy (\%); Bold represents the best test accuracy, and underline represents the runner-up.}
\label{tab:abl_acc}
\end{table}

\end{document}